\documentclass[conference]{IEEEtran}
\IEEEoverridecommandlockouts
\usepackage{cite}
\usepackage{amsmath,amssymb,amsfonts}
\usepackage{algorithmic}
\usepackage{graphicx}
\usepackage{textcomp}
\usepackage{xcolor}
\def\BibTeX{{\rm B\kern-.05em{\sc i\kern-.025em b}\kern-.08em
    T\kern-.1667em\lower.7ex\hbox{E}\kern-.125emX}}
\begin{document}

\title{A Systematic Decade Review of Trip Route Planning with Travel Time Estimation based on User Preferences and Behavior\\
}

\author{\IEEEauthorblockN{Nikil Jayasuriya}
\IEEEauthorblockA{\textit{School of Computing} \\
\textit{Informatics Institute of Technology}\\
Colombo 006, Sri Lanka \\
nikiljay321@gmail.com }
\and
\IEEEauthorblockN{Deshan Sumanathilaka}
\IEEEauthorblockA{\textit{School of Mathematics and Computer Science} \\
\textit{Swansea University}\\
Wales, United Kingdom \\
deshankoshala@gmail.com}

}

\maketitle

\begin{abstract}
This paper systematically explores the advancements in adaptive trip route planning and travel time estimation (TTE) through Artificial Intelligence (AI). With the increasing complexity of urban transportation systems, traditional navigation methods often struggle to accommodate dynamic user preferences, real-time traffic conditions, and scalability requirements. This study explores the contributions of established AI techniques, including Machine Learning (ML), Reinforcement Learning (RL), and Graph Neural Networks (GNNs), alongside emerging methodologies like Meta-Learning, Explainable AI (XAI), Generative AI, and Federated Learning. In addition to highlighting these innovations, the paper identifies critical challenges such as ethical concerns, computational scalability, and effective data integration—that must be addressed to advance the field. The paper concludes with recommendations for leveraging AI to build efficient, transparent, and sustainable navigation systems.
\end{abstract}


\section{Introduction}
Navigation systems have evolved significantly from early cartographic solutions to the sophisticated, real-time route planners we rely on today. With the rise of urbanization and the increasing complexity of transportation networks, modern navigation tools have become integral to our daily lives. Systems like Google Maps, Apple Maps, and Waze are used by millions of people worldwide, providing optimized routing, reducing travel time, and helping individuals navigate complex road networks efficiently. These systems utilize a combination of historical traffic data, GPS signals, and machine learning algorithms to suggest the quickest routes and estimate travel time accurately \cite{b1}, \cite{b2}. However, as urban environments grow increasingly intricate, the limitations of current navigation systems have become more evident.

Despite the widespread use of tools like Google Maps and Waze, current navigation systems are not without their problems. One of the most significant issues is personalization. While these systems offer route suggestions based on general traffic data, they often fail to account for individual user preferences such as avoiding toll roads, preferring scenic routes, or selecting environmentally friendly options. Personalization in route planning is limited, often relying on static user preferences or historical data, which cannot adapt to real-time user behavior or needs \cite{b1}, \cite{b2}. As cities become smarter and transportation networks become more dynamic, the ability of a navigation system to adapt to user preferences in real-time becomes increasingly important \cite{b3}. Moreover, the lack of real-time adaptability is another pressing issue. Current systems provide route recommendations based on historical traffic data, which can be useful for estimating travel time in typical conditions. However, real-time disruptions, such as accidents, road closures, or sudden weather changes, are often not factored into route planning in a sufficiently adaptive manner. Existing systems like Google Maps integrate real-time traffic updates, but these updates are limited in scope and fail to fully account for dynamic, context-dependent changes in user travel patterns, making their recommendations suboptimal in certain situations \cite{b4}, \cite{b5}. The need for systems that can not only predict travel time but also adapt and respond to real-time contextual changes is a critical gap that needs to be addressed.

Furthermore, ethical and privacy concerns have emerged as key issues with the increasing reliance on personal data to improve navigation. Current systems gather a vast amount of data from users to optimize routes, such as location history, travel patterns, and user preferences. However, this data collection often occurs without sufficient transparency, leading to concerns about data privacy and the potential misuse of personal information. Furthermore, algorithmic bias in route recommendations—such as the preferential treatment of toll roads or highways—can lead to inequities in routing suggestions, undermining user trust in the system. Ethical AI practices, including privacy-preserving techniques like federated learning and transparency in algorithmic decision-making, are necessary to address these concerns and ensure the trustworthiness of navigation systems \cite{b6}, \cite{b7}.

Several studies have addressed these issues and proposed solutions, but significant gaps remain. While some studies have focused on improving real-time adaptability and personalization through machine learning techniques, such as collaborative filtering and reinforcement learning, these solutions often lack the ability to continuously learn and adjust to evolving user preferences in real-time \cite{b8}, \cite{b9}. Additionally, despite the growing interest in contextual data integration, many systems still struggle to provide a seamless flow of real-time data, particularly in regions with limited infrastructure \cite{b4}, \cite{b5}. The computational demands of more advanced models, such as reinforcement learning \cite{b9} and graph neural networks \cite{b10}, pose additional challenges for their real-time application in resource-constrained environments like mobile devices. Finally, although much has been written about ethical AI, few studies have explored the practical implementation of privacy-preserving techniques in real-world navigation systems.
Given these gaps, the following research questions arise from the limitations of current systems:

\begin{itemize}
    \item How can navigation systems be made more personalized and adaptable to individual user preferences in real-time?
    \item What are the best methods for integrating diverse real-time contextual data, such as traffic, weather, and user schedules, to improve travel time estimation accuracy and routing efficiency? 
    \item What ethical considerations need to be addressed to ensure that navigation systems are transparent, fair, and respect user privacy
\end{itemize}

In conclusion, the limitations of current navigation systems—such as lack of personalization, limited adaptability to real-time changes, and scalability issues—present significant challenges for their future development. While progress has been made in some areas, such as the integration of real-time traffic data and machine learning for route optimization, gaps remain in providing truly personalized, adaptive, and scalable solutions. By addressing these gaps, future navigation systems can offer a more seamless, efficient, and user-friendly experience, ultimately enhancing urban mobility and contributing to the development of smarter, more sustainable cities.

Moving forward, the paper will discuss the methodology used for the review process, recent works identified during the last decade, limitations and future trends along with the conclusions.

\section{Methodology}

\subsection{Search Strategy}

This systematic review adheres to PRISMA (Preferred Reporting Items for Systematic Reviews and Meta-Analyses) guidelines to ensure a transparent and reproducible methodology for data collection and analysis \cite{b11}. Relevant literature was retrieved from established databases, including IEEE Xplore, SpringerLink, PubMed, Semantic Scholar, ACM Digital Library and Google Scholar, using the following search keywords:

\begin{itemize}
    \item "Business Impact of Navigation Systems"
    \item "Travel Time Estimation"
    \item "Personalized Navigation"
    \item "Adaptive Travel Time Estimation"
    \item "User Preference-Based Navigation"
    \item "AI and Industry 5.0 in Transportation"
\end{itemize}

\begin{figure}[]
    \centering
    \includegraphics[width=\linewidth]{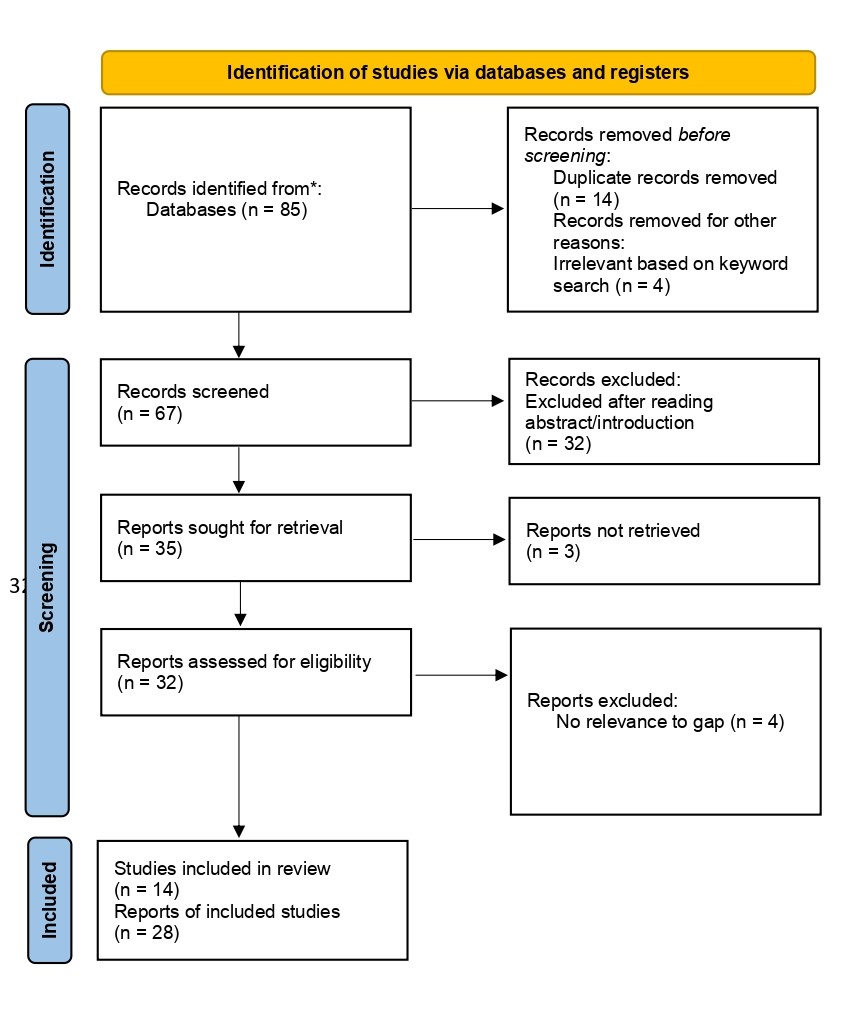} 
    \caption{PRISMA 2020 flow diagram for new systematic reviews which included searches of databases and registers \cite{b12}.}
    \label{fig:prisma_flow}
\end{figure}

To ensure comprehensive coverage, both automated searches and manual screenings of references from selected articles were performed. The PRISMA flow diagram in Figure~\ref{fig:prisma_flow} illustrates the process of identifying, screening, and including studies in the review.

\subsection{ Inclusion and Exclusion Criteria}
\subsubsection{Inclusion Criteria}
\begin{itemize}
    \item Peer-reviewed journal articles, conference proceedings, and industry reports published between 2015 and 2024.
    \item Studies exploring business challenges, customer satisfaction, or operational inefficiencies in navigation systems.
    \item Research discussing Industry 5.0 principles, AI-driven solutions, or collaborative systems for route planning and TTE.
\end{itemize}

\subsubsection{Exclusion Criteria}
\begin{itemize}
    \item Studies solely focused on algorithmic or technical development without addressing business or user-centric contexts.
    \item Non-peer-reviewed articles, including blogs, opinion pieces, and unpublished reports.
\end{itemize}

\subsection{Data Synthesis}
A narrative synthesis was employed to systematically analyze and extract key data points. These data points included:
\begin{itemize}
    \item Business Gaps: Identifying common limitations in existing navigation systems, including unmet personalization needs and operational inefficiencies \cite{b1}.
    \item Market Trends: Highlighting advancements in AI-driven navigation, Industry 5.0 applications, and eco-friendly solutions \cite{b13}.
    \item Future Opportunities: Synthesizing themes to propose directions for research focusing on collaborative transportation networks and smart cities \cite{b14}.
\end{itemize}

The findings were categorized into thematic clusters: unmet user needs, operational challenges, and opportunities for AI-enhanced digital transformation. This approach provides actionable insights into bridging the gaps between technological innovation and business demands.

\section{Recent Works}

This systematic review evaluates a few critical sub-domains within trip route planning and travel time estimation that are directly related to the gaps identified in the project proposal. These gaps include inadequate personalization, limited adaptability to user preferences, and inefficiencies in integrating real-time data for accurate TTE. The sub-domains are analyzed to establish the state of the art and identify areas requiring improvement.

\subsection{Personalization in Navigation}

The inability of existing navigation systems to offer truly personalized routes is a well-documented gap. Tools like Google Maps and Waze provide generalized solutions that fail to account for individual user preferences \cite{b1}, \cite{b2}. Wang et al. \cite{b3} emphasize that the lack of dynamic personalization limits the relevance and satisfaction of route recommendations.

Collaborative filtering techniques, discussed by Li et al. \cite{b8}, offer a framework for tailoring routes by analyzing historical user behavior. However, these models are constrained by their reliance on static data, which does not account for evolving user needs or preferences in real-time. Feedback-based adaptive learning, as explored in Subramaniyaswamy et al. \cite{b3}, presents a promising approach to bridging this gap. It allows systems to update user profiles dynamically, ensuring that routing decisions remain relevant.

The need for multi-objective optimization is another key finding. While most navigation systems focus on minimizing travel time, studies like \cite{b8} propose balancing other factors such as fuel efficiency and toll costs to improve user satisfaction and operational efficiency.

\subsection{Contextual and Dynamic Data Integration}

One of the significant challenges in achieving accurate TTE is the integration of dynamic and contextual data. Some navigation systems often lack the ability to incorporate external factors such as weather, traffic incidents, and user schedules into their routing algorithms \cite{b4}. This limitation is particularly evident in underdeveloped regions, where real-time data coverage is inconsistent.

HERE Technologies’ solutions \cite{b4} demonstrate how real-time traffic data can improve routing, but these systems are often geographically constrained. Deng et al. \cite{b5} suggest that synthetic data generation could address this issue by supplementing datasets in areas with limited coverage. Additionally, event-driven architectures enable navigation systems to respond dynamically to incidents like road closures or accidents \cite{b16}.

Despite these advancements, significant gaps remain in the computational efficiency required to process such large-scale data streams in real-time. Unified data standards and the development of interoperable systems are critical for overcoming these limitations and enhancing the adaptability of navigation tools \cite{b13}.

\subsection{Behavioral Modeling and User Satisfaction}

Behavioral modeling is central to creating a navigation system that learns from user preferences and evolves with their behavior. Subramaniyaswamy et al. \cite{b3} highlight the importance of analyzing long-term user data to improve personalization and relevance. For example, understanding patterns like frequent route deviations or preferred travel times can enable systems to provide proactive recommendations.

\begin{figure}[htbp]
    \centering
    \includegraphics[width=\linewidth]{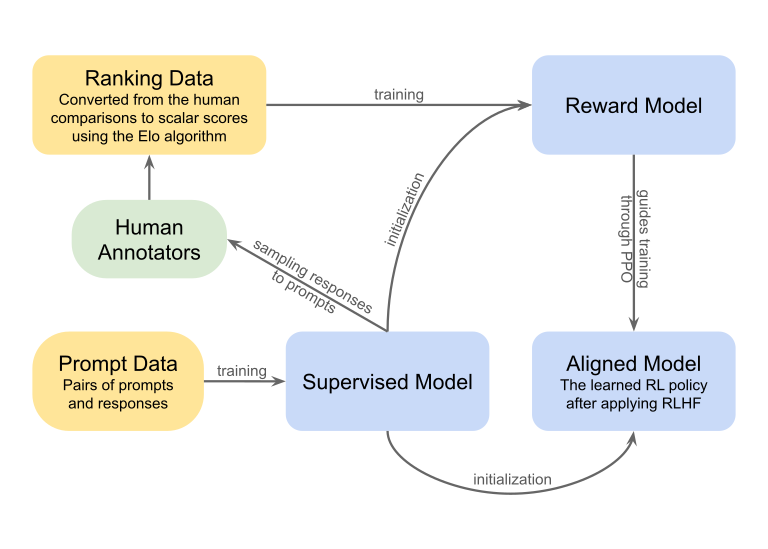}
    \caption{High-level overview of reinforcement learning from human feedback \cite{b17}}
    \label{fig:behavioral_modeling}
\end{figure}

However, the complexity of human decision-making poses a challenge to the effective implementation of behavioral modeling. Li et al. \cite{b8} and Wang et al. \cite{b15} propose integrating machine learning techniques to predict user behavior more accurately, but such models often require significant computational resources and extensive datasets. Sentiment analysis of user feedback, as suggested in \cite{b18}, can complement behavioral analytics by identifying areas for improvement in system design.

Bridging this gap requires systems that combine behavioral insights with real-time adaptability. Such systems would ensure that routing decisions align not only with historical patterns but also with immediate user needs, thereby improving satisfaction and trust.

\subsection{Scalability and Computational Efficiency}

Scalability is a major obstacle to implementing real-time, personalized navigation systems, especially in urban environments with dense traffic. Reinforcement learning (RL)-based approaches, such as those explored by Chen et al. \cite{b9}, offer a dynamic solution by allowing systems to adapt to changing conditions. However, the computational demands of RL models often exceed the capabilities of mobile platforms, making real-time deployment challenging.

Graph Neural Networks (GNNs) have emerged as a scalable alternative, enabling systems to model complex road networks efficiently \cite{b10}. Lightweight versions of GNNs, as proposed by Deng et al. \cite{b5}, are particularly promising for mobile applications. Additionally, edge computing technologies can decentralize data processing, reducing latency and enabling more efficient use of resources \cite{b9}.

To address the scalability gap, future research must focus on optimizing these technologies for real-time operations, ensuring that navigation systems can handle high volumes of data without compromising performance.

\subsection{Ethical and Privacy Considerations}

The reliance on user data to improve personalization and adaptability raises significant ethical and privacy concerns. Tagawa \cite{b7} emphasizes the need for transparent AI models that prioritize data anonymization and user trust. Navigation systems must comply with global privacy standards to ensure that sensitive user information is protected.

Algorithmic biases present another critical challenge. For example, systems that prioritize toll roads or certain routes without clearly communicating these preferences can undermine user trust \cite{b6}. Addressing this issue requires not only technical solutions, such as bias mitigation algorithms, but also regulatory frameworks to govern data usage and decision-making processes.

\subsection{Digital Transformation and Industry 5.0}

The principles of digital transformation and Industry 5.0 present a significant opportunity to bridge gaps in navigation systems, such as limited personalization, inadequate real-time adaptability, and inefficiencies in travel time estimation. By focusing on human-centric AI, collaborative systems, and sustainability, these frameworks enable navigation solutions that address both user and societal needs.

\subsubsection{Redefining Navigation Systems Through Collaboration}

Digital transformation emphasizes creating ecosystems of interconnected systems, where IoT devices, cloud platforms, and edge computing collaborate to provide seamless services. Industry 5.0 enhances this paradigm by prioritizing human-centric design, ensuring technology serves the needs of individuals and communities alike.

In navigation systems, collaboration manifests through integrations with smart city infrastructure, enabling real-time responses to urban dynamics. For instance, HERE Technologies’ real-time traffic management tools \cite{b4} and systems like OpenTripPlanner \cite{b19} demonstrate the feasibility of synchronizing navigation with public transit schedules or urban congestion management policies. These integrations can reduce inefficiencies, offering users multimodal travel options while supporting broader goals such as congestion reduction.

\subsubsection{Human-Centric AI for Navigation}

A fundamental tenet of Industry 5.0 is the use of AI to augment human decision-making, rather than replace it. This aligns directly with the project’s goal of creating adaptive navigation systems tailored to user preferences. Reinforcement learning techniques \cite{b9} can dynamically adapt routing recommendations to evolving user behaviors, while multi-objective optimization models \cite{b8} balance competing priorities like speed, cost, and environmental considerations. For example, route suggestions could prioritize eco-friendly options or incorporate user-specific preferences such as avoiding highways. These features not only enhance usability but also reflect Industry 5.0’s vision of empathetic and responsive systems.

\subsubsection{Sustainability and Resource Optimization}

Digital transformation emphasizes efficiency, and Industry 5.0 expands this by promoting sustainable practices. Navigation systems that incorporate sustainability metrics, such as carbon footprint estimations, can align with eco-conscious user preferences. Masiero et al. \cite{b20} propose routing algorithms that factor in emissions and promote greener travel options, such as carpooling or integrating public transit.

Resource optimization is another critical area. Lightweight AI models and edge computing architectures \cite{b9}, \cite{b5} minimize computational overhead, enabling real-time operations while reducing energy consumption. These solutions align with Industry 5.0’s dual focus on technological advancement and environmental stewardship.

\subsubsection{Expanding the Role of Navigation Systems}

Industry 5.0 reimagines navigation systems as integral components of larger digital ecosystems. Beyond basic routing, these tools can support urban logistics, disaster management, and autonomous vehicle networks. For example, integrating navigation systems with emergency response frameworks could optimize evacuation routes during natural disasters. Similarly, coordination with logistics platforms can streamline last-mile deliveries, reducing delays and costs \cite{b21}.

These expanded applications not only enhance individual user experiences but also contribute to societal resilience, reflecting Industry 5.0’s collaborative ethos.

\section{Limitations and Future Trends}

\subsection{Limitations}

One of the most significant limitations in existing navigation systems is the lack of real-time adaptability to user-specific preferences. Tools like Google Maps and Waze \cite{b1}, \cite{b2} primarily focus on generalized solutions, failing to provide dynamic customization. While collaborative filtering techniques \cite{b8} address historical user data, they often struggle to reflect immediate behavioral changes, leaving a gap in delivering optimal routing.Despite advancements in integrating contextual data such as traffic and weather updates, many systems lack comprehensive and unified data sources. HERE Technologies \cite{b4} and OpenTripPlanner \cite{b19} provide partial solutions but face constraints in underdeveloped regions with poor data infrastructure. This fragmentation limits the effectiveness of real-time travel time estimation, particularly in rural and remote areas. Scalability and computational efficiency remain critical challenges for deploying advanced AI models like reinforcement learning \cite{b9} and Graph Neural Networks \cite{b10} in real-time applications. Mobile devices often lack the processing power required to execute these algorithms without significant latency, affecting usability in high-traffic environments. Navigation systems rely heavily on user data, raising concerns about privacy and algorithmic biases. Bias in routing decisions, such as prioritizing toll roads \cite{b6}, undermines fairness, while inadequate anonymization practices erode trust. Current systems often lack transparent mechanisms to address these issues, highlighting a critical gap in ethical AI deployment \cite{b7}.

\subsection{Future Trends}

The future of navigation systems lies in fully adaptive personalization powered by advanced AI models. Reinforcement learning and multi-objective optimization \cite{b9}, \cite{b8} offer pathways to create systems that dynamically adjust to user preferences while balancing travel time, cost, and environmental impact. By leveraging real-time user feedback, future systems can refine recommendations and ensure continuous improvement. The development of unified data standards and interoperable frameworks can address current fragmentation in contextual data. IoT sensors and smart city integrations \cite{b4} will play a critical role in enabling navigation systems to provide real-time, accurate TTE across diverse regions. Additionally, synthetic data generation \cite{b5} can fill gaps in data-scarce areas, enhancing the reach and reliability of these systems. Future navigation systems must prioritize resource efficiency to ensure scalability. Lightweight versions of GNNs \cite{b10} and edge computing architectures \cite{b9} can decentralize data processing, reducing latency and energy consumption. These advancements will make real-time applications more viable for mobile platforms and resource-constrained environments. The integration of ethical AI practices is essential for building trust and ensuring fairness. Federated learning \cite{b7} can enhance privacy by keeping user data on local devices while enabling collaborative model improvements. Transparent AI models that explain routing decisions can mitigate biases and enhance user trust, aligning with the principles of Industry 5.0. Sustainability will be a cornerstone of future navigation systems, aligning with global efforts to reduce emissions and promote eco-friendly mobility. Routing algorithms that incorporate carbon footprint metrics [19] and integrate public transit options \cite{b19} can support greener travel. Furthermore, navigation systems can expand their role by integrating with logistics platforms and emergency response frameworks, enhancing their societal impact.

\begin{table*}[htbp]
\centering
\caption{Bridging Limitations with Future Trends in Navigation Systems}
\begin{tabular}{|p{2cm}|p{6cm}|p{6cm}|p{2cm}|}
\hline
\textbf{Limitation}                     & \textbf{Description}                                                                                         & \textbf{Future Trend}                                                                 & \textbf{References}         \\ \hline
Lack of real-time personalization       & Navigation systems fail to dynamically adapt to evolving user preferences and behaviors.                      & Adaptive personalization using reinforcement learning and real-time feedback loops.   & \cite{b3}, \cite{b8}, \cite{b9} \\ \hline
Fragmented contextual data              & Data from traffic, weather, and user-specific contexts remain disjointed, especially in underserved areas.     & Unified ecosystems with IoT-based integration and synthetic data generation.         & \cite{b4}, \cite{b5}, \cite{b13}\\ \hline
Privacy and bias concerns               & Overreliance on sensitive user data raises trust issues and risks algorithmic bias.                           & Ethical AI frameworks, federated learning, and transparent algorithms.               & \cite{b6}, \cite{b7}, \cite{b20}             \\ \hline
Limited sustainability considerations   & Current systems neglect environmental impacts in routing decisions.                                           & Routing with carbon footprint metrics and integration of public transit options.     & \cite{b13}, \cite{b19}                             \\ \hline
\end{tabular}
\label{tab:limitations_future_trends}
\end{table*}

\section{Conclusion}

This paper systematically reviewed the advancements, limitations, and future trends in trip route planning and travel time estimation, emphasizing gaps in personalization, adaptability, and computational efficiency. By leveraging the principles of Industry 5.0 and digital transformation, the project aims to address these deficiencies, creating a navigation system that dynamically adapts to user preferences while ensuring scalability and ethical compliance. The following subsections summarize key findings, highlight relevant datasets for future development, and propose evaluation metrics to assess system effectiveness.

\subsection{Summary of Findings}

The review revealed significant gaps in current navigation systems. Personalization remains limited, with tools like Google Maps and Waze relying on generalized solutions that fail to dynamically adapt to evolving user preferences \cite{b1}, \cite{b2}. Although collaborative filtering techniques and reinforcement learning approaches \cite{b8}, \cite{b9} offer potential solutions, they often lack the real-time feedback necessary to reflect immediate user behavior. Fragmented contextual data further exacerbates the problem, with rural and underrepresented areas particularly affected by inconsistent data availability. Real-time traffic solutions like those provided by HERE Technologies \cite{b4} and OpenTripPlanner \cite{b19} demonstrate advancements, but the lack of unified data standards hinders seamless integration. Computational inefficiencies also persist, with resource-intensive AI models like Graph Neural Networks (GNNs) and reinforcement learning posing challenges for deployment on mobile and edge devices \cite{b9}, \cite{b10}. Finally, ethical and privacy concerns, including biases in routing algorithms and inadequate anonymization practices, undermine user trust in current systems \cite{b6}, \cite{b7}. Addressing these gaps is essential for developing navigation systems that are not only accurate and efficient but also user-centric and trustworthy.

\subsection{Relevant Databases and Data Sources}

Developing and evaluating the proposed system requires diverse datasets that improve real-time travel time estimation and enable personalization. Key sources identified in this review include traffic and transit data, synthetic data, behavioral data, and weather data. These datasets are essential, but they also present limitations.

A significant challenge is data gaps and incomplete coverage, particularly in rural areas where real-time traffic data is sparse. Systems like HERE Technologies \cite{b4} and OpenTripPlanner \cite{b19} provide comprehensive coverage in urban areas, but rural regions often lack sufficient data, leading to less accurate routing. Furthermore, data accuracy and quality is a concern. Synthetic data, while useful for supplementing real-world data \cite{b5}, may not fully capture the complexities of traffic conditions or user behavior. For example, simulated data may fail to account for sudden incidents like accidents or weather disruptions, leading to inaccuracies in travel time predictions and routing suggestions.

Addressing these challenges requires better integration of diverse data sources and improvements in synthetic data models to ensure more accurate and dynamic routing decisions.

\subsection{Evaluation Metrics}

To measure the system’s effectiveness, clear and quantifiable evaluation metrics are essential. Travel time estimation accuracy will be assessed using Mean Absolute Error (MAE) to ensure close alignment between estimated and actual travel times, aiming for an error margin of less than 5\% across diverse scenarios. Personalization effectiveness can be measured by the percentage of user-suggested routes adopted by the system over time, with a target of achieving over 90\% alignment within three months of deployment. Scalability and computational efficiency will be evaluated by measuring system latency under high traffic densities, with a goal of sub-second response times for 95\% of requests. The system’s adaptability to contextual data, such as traffic and weather updates, will be evaluated by its ability to make accurate, real-time routing adjustments, targeting over 85\% accuracy in dynamic environments. Ethical compliance is equally critical, with metrics focusing on the percentage of anonymized user data and the absence of algorithmic biases in route suggestions. These metrics ensure that the system not only addresses technical gaps but also aligns with user expectations for trustworthiness and fairness.

\subsection{Remarks}

This study highlights the transformative potential of aligning navigation systems with Industry 5.0 principles and digital transformation goals. By prioritizing personalization, scalability, and ethical AI, the proposed system can redefine the navigation landscape, improving user satisfaction and operational efficiency. Future work should focus on integrating the identified datasets, refining adaptive AI models, and conducting real-world trials to validate the proposed evaluation metrics.

\vspace{12pt}

\end{document}